\documentclass[letterpaper, 10 pt, conference]{ieeeconf}  

\newcommand{\shrinka}{\def\baselinestretch{0.993}\large\normalsize}
\usepackage{graphicx}
\usepackage{amsfonts} 
\usepackage{cite}
\graphicspath{{./images/}}
\usepackage{amsmath}

\DeclareMathOperator*{\argmin}{arg\,min}
\usepackage[skip=0pt,font=small,labelfont=bf]{caption}

\IEEEoverridecommandlockouts                              

\title{\LARGE \bf
23 DoF Grasping Policies from a Raw Point Cloud
}

\author{Martin Matak$^{1}$, Karl Van Wyk$^{2}$, Tucker Hermans$^{1,2}$
\thanks{$^{1}$School of Computing and the Robotics Center, University of Utah.       
$^{2}$NVIDIA, USA. {\tt\small martin.matak@utah.edu}}}%

\usepackage{color}
\definecolor{international_orange}{RGB}{240, 74, 0}
\definecolor{pacific_blue}{RGB}{2,108,181}

\begin{document}

\input{intro-fig}
\maketitle

\begin{abstract}
Coordinating the motion of robots with high degrees of freedom (DoF) to grasp objects gives rise to many challenges. In this paper, we propose a novel imitation learning approach to learn a policy that directly predicts 23 DoF grasp trajectories from a partial point cloud provided by a single, fixed camera. At the core of the approach is a second-order geometric-based model of behavioral dynamics. This Neural Geometric Fabric (NGF) policy predicts accelerations directly in joint space. We show that our policy is capable of generalizing to novel objects, and combine our policy with a geometric fabric motion planner in a loop to generate stable grasping trajectories. We evaluate our approach on a set of three different objects, compare different policy structures, and run ablation studies to understand the importance of different object encodings for policy learning.
\end{abstract}

\section{Introduction}

Autonomous grasping with multi-fingered hands has the potential to enable high throughput, generalized pick-place operations across a wide range of objects. However, achieving high performance in this domain is difficult due to the high-dimensional embodiments, sensory noise, partial observability, and large variation in object dynamics. Existing strategies for computing grasping trajectories are typically an outcome of a complex optimization on multiple levels of the stack ~\cite{ciocarlie-ijrr2009-eigengrasp, chen-2018-pSDF, miller-icra99-force-wrench-space, suarez-2009tro-contact-regions, suarez-2011ijrr-synthesizing-grasps, zheng-2005ijrr-force-closure-uncertainty, hang-tro2016-hierarchical-fingertip-space, siddiqui-frontiers2021-bayesian-exploration, lu-isrr2017-grasp-inference, lu-ral2019-grasp-type, lu-ram2020-MultiFingeredGP,lu-iros2020-active-grasp,vandermerwe-icra2020-reconstruction-grasping, matak-ral23-precision-grasps, mousavian-iccv2019-graspnet}. While optimization-based approaches might lead to high grasp success rates, they suffer from being computationally expensive, especially in these high-dimensional, continuous action spaces. 

One way to overcome this limitation is the use of geometrically aware policy structures for control in high-dimensional continuous spaces. Such approaches include Dynamic Movement Primitives (DMPs) \cite{ijspeert2013dynamical}, Riemannian Motion Policies (RMPs) \cite{ratliff2018riemannian}, Geometric Dynamics Systems (a stable subclass of RMPs) \cite{cheng2020rmp}, Operational Space Control (OSC) \cite{khatib1987unified}, and geometric fabrics \cite{karl-ral22-fabrics}. Geometric fabrics are nonlinear, second-order differential equations that are provably stable and have been shown to outperform these existing control methods in learning contexts \cite{mandy-ngf-corl22}. While fabrics have been used for grasping~\cite{mandy-ngf-corl22,chen-implicit-shape-corl22}, they have not yet been shown to generalize to novel object geometries from a single camera view. Xie et al.~\cite{mandy-ngf-corl22} leverage estimates of object pose to train object-specific grasping policies that mimic human demonstrations. Chen et al.~\cite{chen-implicit-shape-corl22} leverage a manually derived geometric fabric to move the robot to a single predicted grasp palm pose and hand configuration. 

The main contribution of this paper is a model that reliably predicts a smooth 23-DoF grasping trajectory directly in joint space to grasp a previously unseen object given only a single RGBD camera view, when the robot is placed in the general vicinity (not a specific grasp target) of the object for grasping. We create an NGF policy to learn this behavior, building upon the prior success of NGFs~\cite{mandy-ngf-corl22}. An example trajectory is shown in Fig.~1. Our results show that NGFs produce smooth and stable behavior that generalizes over object shape and pose variation. An off-the-shelf motion planner drives the robot to the general vicinity of the object, alleviating the burden of learning how to grasp from arbitrary far away, initial robot configurations.


\section{Imitation Learning Setup}
We aim to learn policies that generate trajectories to grasp objects of interest given the current state of the robot, an object encoding \textbf{o}, and an initial robot placement in the vicinity of the object. We train these policies via imitation learning by first generating a dataset of successful grasping trajectories in simulation using geometric fabrics \cite{karl-ral22-fabrics} and subsequently construct a surrogate expert as a function of the data. This allows for on-policy imitation learning, DAgger \cite{ross-2011-dagger}, to train our NGF policy.

\subsection{Data Collection}
Data collection consists of two primary steps: (1) finding grasps and (2) generating robust grasping trajectories. 

\textbf{Finding in-contact grasps} We place an object onto a table and try to find a grasp that lifts the object successfully. To do so,  we use a dataset of grasps collected by~\cite{dylan-icra2023-dexgrasp}. After filtering out the poses for reachability and collision, we move the robot's EE to the remaining poses, one by one, until a successful grasp is found or all poses are exhausted. The preshape hand configuration is fixed and a simple heuristic is used to close the hand. If the grasp is successful, we store it in a set of successful grasps. We repeat this process for different objects and poses.

Next, we cluster the successful grasps based on EE pose in the object frame and save only $N$ grasp poses per object ($N=7$). We do this step hoping to have a low variance in grasp distribution, which eases the learning process. Then, we place the object across the grid on the table and try out the $N$ grasps one by one until we find a successful one, which we store in the dataset $\mathcal{D}_g$ we use to generate training trajectories. Here $\mathcal{D}_g=\{(\mathbf{q}^g,\textbf{o})_i\}_{i=1}^{i=|\mathcal{G}|\times|\mathcal{O}|}$ where $\mathcal{G}$ is set of poses on the grid, $\mathcal{O}$ is set of objects, $\mathbf{q}^g$ is robot joint position and \textbf{o} is object encoding. 

\begin{figure}
  \includegraphics[width=0.22\textwidth]{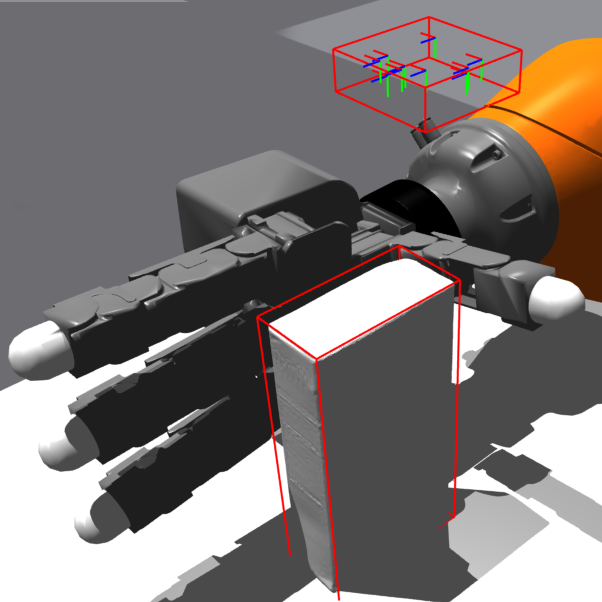}
  \includegraphics[width=0.22\textwidth]{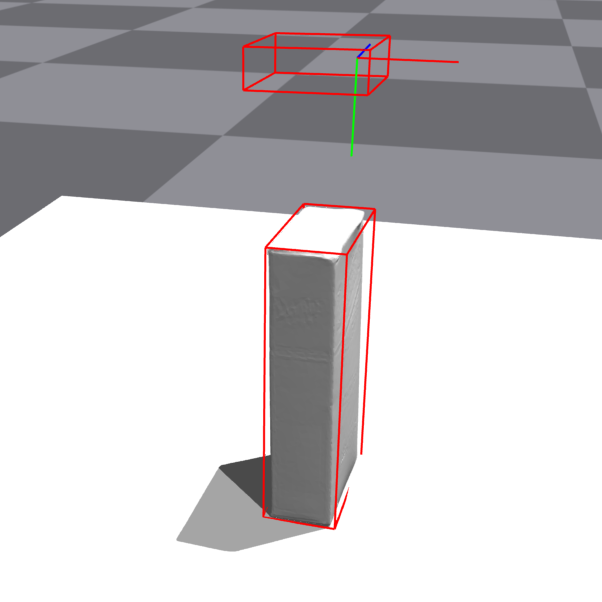}
  \caption{Region which we sample from above the object is computed based on the object's bounding box. \textbf{Left: }Training data is collected by generating trajectories from the grasp configuration to the sampled poses and then reversing those trajectories. \textbf{Right:} At evaluation time, we sample a target pose above the object and use a motion planner to get there. Then, the policy predicts a trajectory to grasp the object.}
  \label{fig:point-above-object}
\end{figure}

\textbf{Generating trajectories} We initialize the scene in a configuration $(\mathbf{q}^g,\textbf{o})_i$ from the dataset and sample a target pose $\mathbf{x_0}$ for the EE above the object in a prespecified region, as shown in Fig.~\ref{fig:point-above-object}. This region is a design choice and any other region could be used instead. Then, we use geometric fabrics \cite{karl-ral22-fabrics} as a motion planner to move the robot from the grasp configuration $\mathbf{q}^g$ to the target pose $\mathbf{x_0}$, resulting in a trajectory $\overleftarrow{\tau_i} = [(\mathbf{q}^g, \dot{\mathbf{q}}^g), ..., (\mathbf{q}^0, \dot{\mathbf{q}}^0)]$. Importantly, here we add the object as an obstacle when computing a trajectory to the target pose to ensure collision free trajectory. Finally, we reverse the generated trajectory to store a trajectory that moves into the grasp configuration, $\overrightarrow{\tau_i} = \texttt{reverse}( \overleftarrow{\tau_i})$. We generate $M$ trajectories per object pose ($M=256$). We further encode the object's partial view point cloud $\mathbf{z}=F_\theta(\textbf{o})$ where $F_\theta$ is the encoder described in Section \ref{subsec:object_encoding}. This results in a dataset $\mathcal{D}_\tau=\{(\overrightarrow{\tau_m},\mathbf{z}_i)\}_{i=1,m=1}^{i=K,m=M}$ where $|\mathcal{D}_\tau| = MK$ that we use for training our policies using imitation learning.

\subsection{Surrogate Expert and Training}\label{sec:expert}
With the above dataset $\mathcal{D}_\tau$, we now construct a surrogate expert that can be efficiently queried on-the-fly, enabling DAgger-style imitation learning. First, we select a randomly sampled configuration, $\mathbf{q}_t^d, \dot{\mathbf{q}}_t^d$, from trajectory $\overrightarrow{\tau_k}$ at time index $t$ in our dataset. Then, the surrogate expert, $\pi_e(\cdot)$, uses a PD controller to compute an acceleration action to steer towards the next configuration along the trajectory, $(\mathbf{q}_{t+1}^d, \dot{\mathbf{q}}_{t+1}^d) \in \overrightarrow{\tau_k}$. With a sufficient number of consecutive executions, $\pi_e$ will produce motion that collapses onto trajectory $\overrightarrow{\tau_k}$ from the dataset. Formally, this expert is: 
\begin{align}
    \pi_e(\mathbf{q}_{t+1}^d, \mathbf{q}_t, \dot{\mathbf{q}}_t) = k_p (\mathbf{q}_{t+1}^d - \mathbf{q}_t) - k_d \dot{\mathbf{q}_t}
\end{align}
where $\mathbf{q}_t$ and $\dot{\mathbf{q}}_t$ are the current joint position and velocity. With this expert in place, we train  policy $\pi_{\theta}$ online to predict desired accelerations following the optimization problem 
\begin{align}
\theta^* = \argmin_{\theta} \quad & || \pi_{\theta}(\mathbf{q}_t, \dot{\mathbf{q}}_t) - \pi_e(\mathbf{q}_t^d, \mathbf{q}_t, \dot{\mathbf{q}}_t)||
\end{align}
where $||\cdot||$ is L2 norm and this loss function is calculated over a batch of evaluations before updating the policy parameters. In practice, after sampling $\mathbf{q}_t^d, \dot{\mathbf{q}}_t^d$, we add $\epsilon$ noise from a narrow uniform distribution to obtain $\mathbf{q}_t, \dot{\mathbf{q}}_t$. We then identify the trajectory $\overrightarrow{\tau_g} \in \mathcal{D_\tau}$ that has the closest configuration $\mathbf{q}_i^d, \dot{\mathbf{q}}_i^d$ ($i$ is some time index) to $\mathbf{q}_t, \dot{\mathbf{q}}_t$. The expert $\pi_e$ then targets subsequent configurations in this identified trajectory $\overrightarrow{\tau_g}$ for the remainder of the rollout of the policy. 

\section{Neural Policy Architectures}
In this section, we provide a short introduction to geometric fabrics \cite{karl-ral22-fabrics} and describe our instantiated NGF policy. We additionally cover the baseline policy, discuss object encodings, numerical integrators for our policies, and connect our policies with off-the-shelf motion planners.

\subsection{Geometric Fabrics}
Geometric fabrics are a specific class of autonomous, second-order differential equations that are provably stable (i.e., they are guaranteed to come to rest at a local minimum) and exhibit path consistency in the motion they generate due to their geometric nature. They consist of three major components: 1) an energized geometry, 2) a driving force that derives from a scalar potential function, and 3) damping. Geometric fabrics are dependent on joint position $\mathbf{q}$, joint velocity $\dot{\mathbf{q}}$, and an additional feature vector $\mathbf{z}$, that together form the following system,
\begin{align} \label{eq:GeneralSystemForm}
    \nonumber
    \ddot{\mathbf{q}} =& \widetilde{\mathbf{h}}(\mathbf{q}, \dot{\mathbf{q}}, \mathbf{z} ) + \alpha_{L} \dot{\mathbf{q}} - \mathbf{M}^{-1} (\mathbf{q}, \hat{\dot{\mathbf{q}}}, \mathbf{z} ) \partial_\mathbf{q} \psi (\mathbf{q}, \mathbf{z} ) \\
    & -\mathbf{B}(\mathbf{q}, \dot{\mathbf{q}}, \mathbf{z}) \dot{\mathbf{q}} - \beta \dot{\mathbf{q}},
\end{align}
where $\widetilde{\mathbf{h}}$ is a nonlinear geometry (homogeneous of degree 2 in velocity), $\alpha_L$ is an energization coefficient calculated to conserve some particular energy, $\mathbf{M}^{-1} \partial_\mathbf{q} \psi$ creates the driving force and dictates the local minima of the fabric, $\mathbf{B}$ is a postive semi-definite damping matrix, $\hat{.}$ denotes a unit vector, and $\beta \in \mathbb{R}^+$ is an additional damping scalar. Full description of these components are outside the scope of this paper and we refer the reader to \cite{nathan-icra21-finsler} for in-depth discussion on generalized nonlinear geometries and \cite{karl-ral22-fabrics} for geometric fabrics and relevant tools like the energization operator.

\subsection{Neural Geometric Fabric}
We instantiate a Neural Geometric Fabric by parameterizing the various components of a geometric fabric via neural networks and object encoding as the feature vector $\mathbf{z}$. For the following, $\mathbf{F}_\theta (\cdot)$ is a neural network consisting of three fully connected, feedforward layers with 512 units each and with hidden layer ELU activations, $\theta$ trainable parameters, and 1000 Random Fourier Features (RFFs) over its inputs. Every usage of $\mathbf{F}_\theta(\cdot)$ below is meant as a different neural network instance. For some outputs like the damping coefficient below, the final output layer is also passed through ELU activations to force positive outputs. Otherwise, the final layer is linear.

The geometric portion of the fabric is produced by the metric, $\mathbf{M}_g = \mathbf{U}_g\mathbf{U}_g^T$, with $\mathbf{U}_g = \mathbf{F}_g (\mathbf{q}, \hat{\dot{\mathbf{q}}}, \mathbf{z})$, a lower triangular matrix with positive diagonal elements. The geometric acceleration results from $\pi_g = \dot{\mathbf{q}}^T \dot{\mathbf{q}} \: \mathbf{F}_{\mathcal{X}} (\mathbf{q}, \hat{\dot{\mathbf{q}}}, \mathbf{z})$. The final geometric force is then $\mathbf{f}_g = \mathbf{M}_g \pi_g$.

The driving force of the fabric is produced by the metric, $\mathbf{M}_f = \mathbf{U}_f\mathbf{U}_f^T$, with $\mathbf{U}_f = \mathbf{F}_f (\mathbf{q}, \mathbf{z})$, a lower triangular matrix with positive diagonal elements. The associated scalar potential force results from taking the gradient $\partial_\mathbf{q} \psi$ of scalar function $\psi = \mathbf{F}_\psi (\mathbf{q}, \mathbf{z})$. This force is coupled with a learned positive damping scalar, $\beta_f = \mathbf{F}_\beta(\mathbf{q}, \dot{\mathbf{q}}, \mathbf{z})$. The final driving force is then $\mathbf{f}_f = \partial_\mathbf{q} \psi + \beta_f \dot{\mathbf{q}}$.

The metrics and forces are summed to produce 
\begin{align}
    (\mathbf{M}_g + \mathbf{M}_f) \ddot{\mathbf{q}} + \mathbf{f}_g + \mathbf{f}_f = \mathbf{0},
\end{align}
which can be resolved as
\begin{align}
    \ddot{\mathbf{q}} = \widetilde{\mathbf{h}} - \mathbf{M}^{-1} \partial_\mathbf{q} \psi -\mathbf{B} \dot{\mathbf{q}}
\end{align}
where $\mathbf{M} = \mathbf{M}_g + \mathbf{M}_f$, $\widetilde{\mathbf{h}} = - \mathbf{M}^{-1} \mathbf{f}_g$, and $\mathbf{B} = \beta_f \mathbf{M}^{-1} \dot{\mathbf{q}}$. We calculate an energization coefficient, $\alpha_L$, using the energy $L = \frac{1}{2} \dot{\mathbf{q}}^T \dot{\mathbf{q}}$ and geometric acceleration $\ddot{\mathbf{q}} = \widetilde{\mathbf{h}}$, and add this additional acceleration as $ \alpha_{L} \dot{\mathbf{q}}$. Finally, we add an additional positive scalar damping with $\beta = 5$ resulting in the final fabric form in (\ref{eq:GeneralSystemForm}). Practically, this additional damping helped stabilize training.


\subsection{Acceleration MLP}
We implement an MLP-based acceleration policy as a baseline. Input and output of the policy is the same as for NGF, i.e. input joint position, velocity, and object encoding and output joint acceleration. The input layer is followed by two hidden layers before the output layer. The first hidden layer has a dimensionality of 64, while the second one has 256. The ReLU activation function is used after each layer.

\subsection{Object encoding}
\label{subsec:object_encoding}
\begin{figure}
  \includegraphics[width=0.22\textwidth]{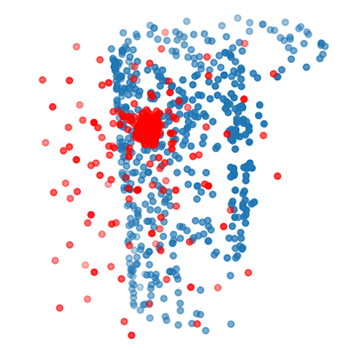}
  \includegraphics[width=0.22\textwidth]{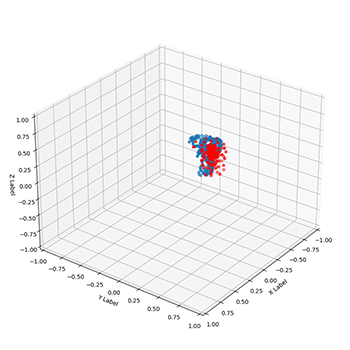}
  \caption{Example point cloud reconstruction; input in blue, output in red. While imperfect (left), it preserves information about object pose and course geometry (right).}
  \label{fig:reconstruction}
\end{figure}

We aim to generate grasp trajectories that generalize across poses and object shapes directly from a point cloud. Thus, we need to encode the object's shape and pose. We do so by training an autoencoder on a reconstruction task for pointclouds stored in robot base frame and then use only the encoder from the autoencoder to generate an object encoding. After training on the reconstruction task, the encoder is not further modified, i.e., the weights are frozen. The loss function for the reconstruction task is a weighted sum of Chamfer distance and L2 regularization on the latent space. The data for the reconstruction task is generated by placing an object at 9k different poses across the table and saving the point cloud obtained from the fixed camera. We repeat this process for 4 YCB~\cite{ycb-dataset} objects, producing in total 36k samples. The encoder is based on PointNet layers~\cite{pointnet-cvpr2017} from PyTorch Geometric library~\cite{pytorch-geometric} while the decoder is a 3 layer MLP. An example of a reconstructed point cloud can be seen in Fig.~\ref{fig:reconstruction}. While imperfect, we show in our experiments that it is sufficiently good for generating grasp trajectories. 

\subsection{Acceleration Integrator}
Both the learned policies and the surrogate expert described in Sec.~\ref{sec:expert} produce accelerations. We integrate these accelerations forward in time via the approximate RK2 integration scheme as described in~\cite{gruver2022deconstructing}. The exact formulation calculates the next joint position and velocity, $\mathbf{q}_{t+1}$ and $\dot{\mathbf{q}}_{t+1}$, from the current joint position and velocity, $\mathbf{q}_t$ and $\dot{\mathbf{q}}_t$, acceleration $\ddot{\mathbf{q}}_t$, and timestep $\Delta t$ as
\begin{flalign}
\mathbf{q}_{t+1} &= \mathbf{q}_t + \Delta t \dot{\mathbf{q}}_t + \frac{1}{2} \Delta t ^2 \ddot{\mathbf{q}}_t \\
\dot{\mathbf{q}}_{t+1} &= \dot{\mathbf{q}}_t + \Delta t \ddot{\mathbf{q}}_t
\end{flalign}
The policies perform control at 30 Hz and therefore $\Delta t = \frac{1}{30}$.

 \subsection{Motion Planner In The Loop}
 A further shortcoming of the previous work \cite{mandy-ngf-corl22} is a necessity to always start from a precise initial configuration. While the ability to learn longer trajectories showcases the model's capabilities, we believe one should use existing (traditional) tools when a solution is well known and use learning models only when necessary. To underline this, we leverage an existing off-the-shelf motion planner to move close to the object and lift it after the grasping is completed. The learned model learns only the complex component - grasping the object. 
 
 When deploying our model, we sample a pose above the object (Fig.~\ref{fig:point-above-object}) in the same way as we do when generating the trajectories during training. Then, instead of solving for IK, we look for the closest target configuration in the training dataset. To find such a configuration, we find an encoding in the dataset that is closest (L2) to the encoding of the current object we want to grasp. Then, we search across trajectories associated with that encoding and find the one whose EE pose starts closest to the queried pose above the object. We set the corresponding configuration as the target configuration for the motion planner. We found that if we would naively solve IK instead, the deployed policy wouldn't perform as well due to the out-of-distribution robot configurations. We plan on learning this as a separate model for future work since this querying can be expensive if the whole dataset can't fit onto a GPU.

\section{Experiments}
\begin{figure}[htbp]
  \centering
  \includegraphics[width=0.465\textwidth, trim={0cm 10cm 4cm 1cm},clip]{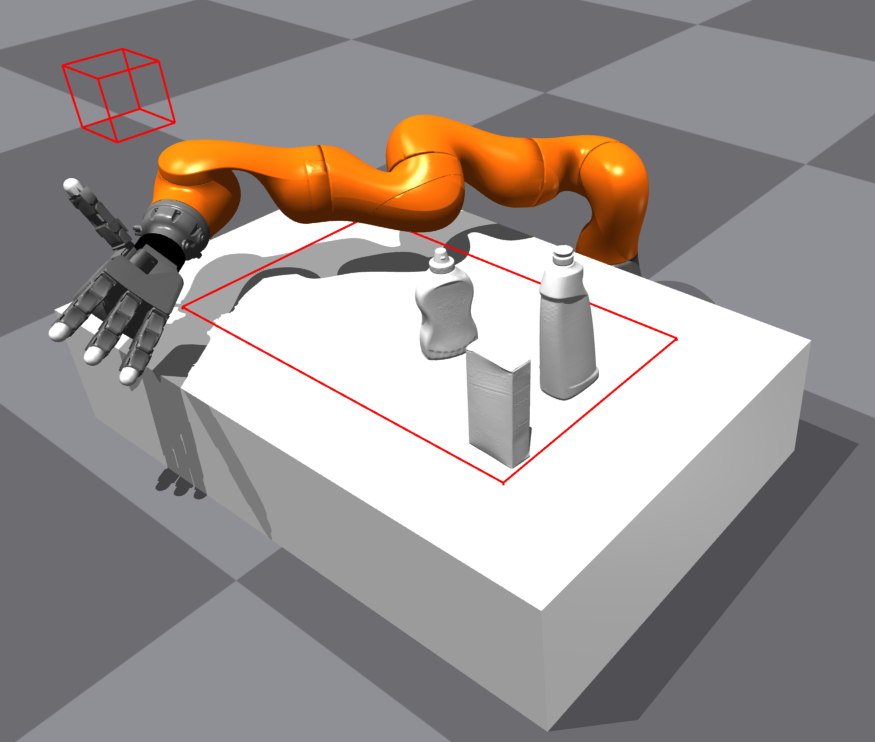}
  \includegraphics[width=0.15\textwidth]{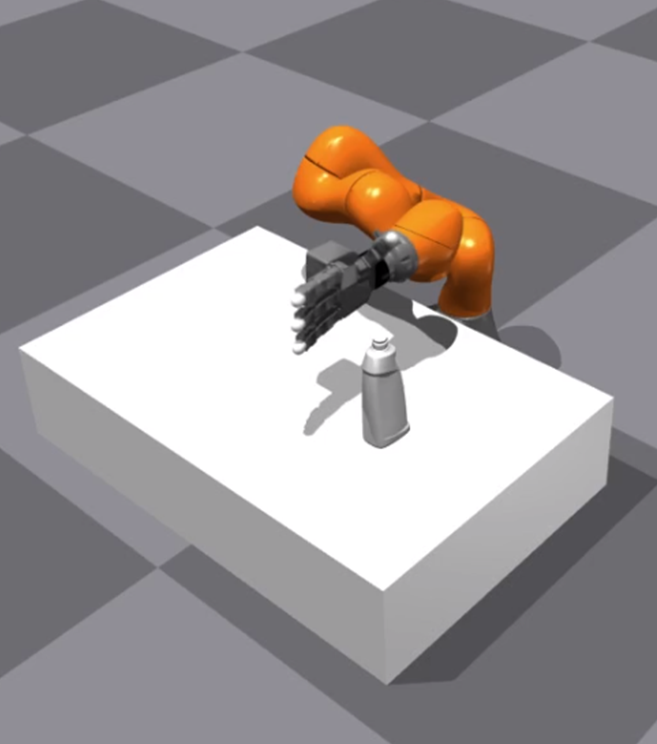}
  \includegraphics[width=0.15\textwidth]{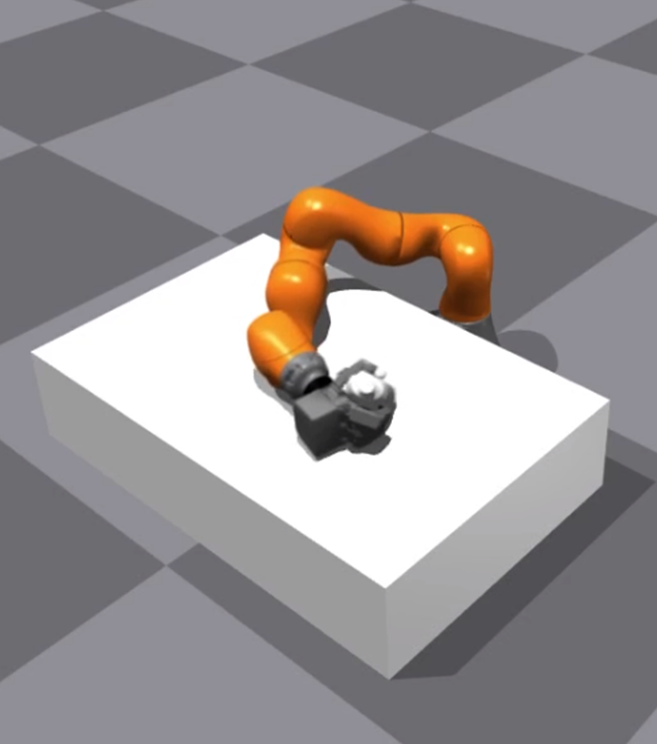}
  \includegraphics[width=0.15\textwidth]{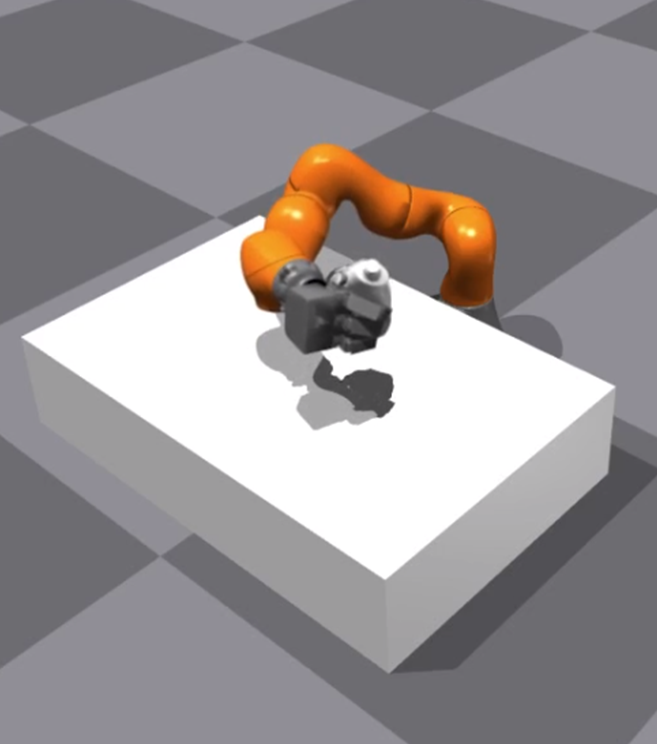}
  \caption{\textbf{Top}: The red lines show the camera location and workspace boundaries. The objects are (left to right): `mustard', `sugarbox', `bleach'. \textbf{Bottom}: Our manipulation pipeline from left to right: a motion planner moves the robot close to the object, our policy grasps the object, and then a motion planner lifts the object.}
  \label{fig:pipeline}\vspace{-12pt}
\end{figure}

 We place an object at a random position on the table and simply fit a spline as a motion planner to get from a fixed initial configuration close to the object. After that, we execute a grasping policy to grasp the object. Finally, we use a motion planner \cite{karl-ral22-fabrics} to lift the object. If the robot holds the object after executing the lifting trajectory, we label the attempt as a successful grasp. We repeat this 100 times per object. Our experimental setup is shown in Fig.~\ref{fig:pipeline}.

\begin{figure}
\centering
\includegraphics[width=\linewidth]{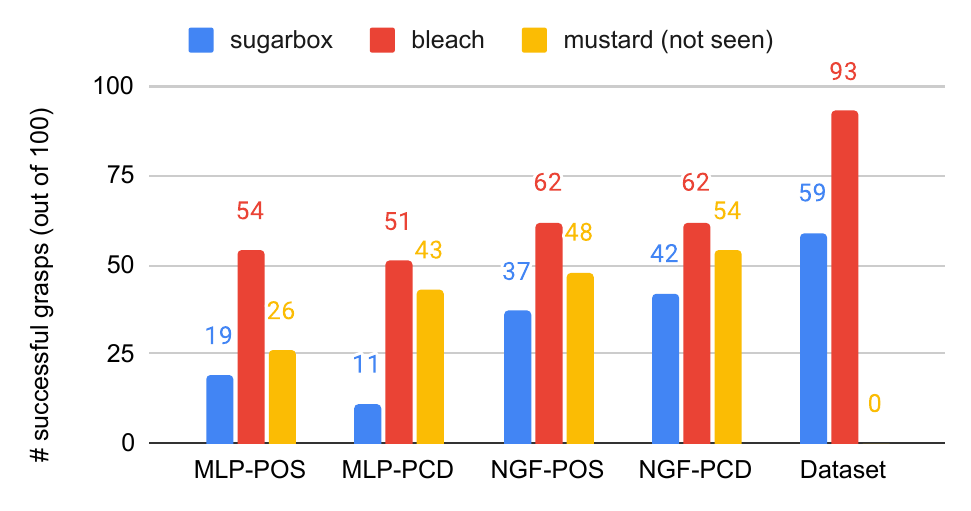}
\caption{Grasp success rates for different grasp-trajectory predicting models and the dataset we use for training. Not all samples in the dataset are successful grasps. Input modalities: POS-position only, PCD - pointcloud.}
\label{fig:plot_success} \vspace{-18pt}
\end{figure}

 We evaluate our approach across 3 different objects. The grasp policies have been trained on grasp trajectories for 2 (bleach, sugarbox) out of the 3 objects. Results are shown in Fig.~\ref{fig:plot_success} with discussions below. We note that when we ran the trajectories from the dataset in the simulator, only $59\%$ of trajectories successfully picked up `sugarbox` while that number is $93\%$ for `bleach.` We think it is important to keep these numbers in mind when reading the results in Fig.~\ref{fig:plot_success}.


\noindent \textbf{Effect of Policy Structure:} When presented with the same object encoding, NGF always outperforms MLP policy across all objects. Most importantly, when a model is presented with an encoded pointcloud, NGF outperforms MLP.  

\noindent \textbf{Object Representation:} We investigate the impact of object encoding on policy performance across two cases: 1) passing the position (-POS) of the object to the policy, and 2) passing the latent point cloud encoding (-PCD) to the policy. Overall, NGF performance with object position is fairly competitive to the MLP baseline. However, the PCD encoding does improve the NGF policy performance over the position only input. These results reveal that the latent encoder for the \textit{partial} view point cloud produces useful features for grasping operations that meet or exceed position-only policy performance. The main reason for this difference is that the point cloud contains information about the geometry of the object while the position only input does not. This additional information is useful for informing grasp behavior. Overall, this result is encouraging and suggests that effective high-DoF, high-frequency policies can be trained that leverage camera views, which contain only partial information about the scene. This capability will enable policies to be immediately more useful for real-world deployment. 



\section{Conclusion}
 We show how to train an NGF policy without any human demonstrations and how to use the policy in a loop with an off-the-shelf motion planner. Our policy generates 23 DoF grasp trajectories that generalize across previously unseen objects directly from a partial view point cloud.

\addtolength{\textheight}{-7cm}   




\section*{ACKNOWLEDGMENT}
This work was supported in part by NSF Award \#1846341.
\newpage\clearpage

\bibliographystyle{IEEEtran}
{
\footnotesize
\bibliography{references}
}

\end{document}